\documentclass{sig-alternate-05-2015}

\usepackage[algo2e, noend, noline]{algorithm2e}
\usepackage{subcaption}

\usepackage{tcolorbox}

\definecolor{darkred}{RGB}{193,81,109}
\definecolor{lightblue}{RGB}{169,229,226}
\definecolor{darkblue}{RGB}{64,126,203}
\definecolor{purple}{RGB}{131,92,187}

\definecolor{fig2_red}{RGB}{239,0,0}
\definecolor{fig2_green}{RGB}{118,199,121}
\definecolor{fig2_blue}{RGB}{138,199,234}
\definecolor{fig2_purple}{RGB}{149,122,197}

\providecommand{\DontPrintSemicolon}{\dontprintsemicolon}
\DontPrintSemicolon


\begin{document}




\clubpenalty=10000
\widowpenalty = 10000
\title{PathNet: Evolution Channels Gradient Descent in Super Neural Networks}
\numberofauthors{1}
\author{
\alignauthor
Chrisantha Fernando, Dylan Banarse, Charles Blundell, Yori Zwols, David Ha$^\dagger$, Andrei A. Rusu, Alexander Pritzel, Daan Wierstra\\
       \affaddr{Google DeepMind, London, UK. $^\dagger$Google Brain}\\
       \email{chrisantha@google.com}
}

\maketitle
\begin{abstract}
For artificial general intelligence (AGI) it would be efficient if multiple users trained the same giant neural network, permitting parameter reuse, without catastrophic forgetting. PathNet is a first step in this direction. It is a neural network algorithm that uses agents embedded in the neural network whose task is to discover which parts of the network to re-use for new tasks. Agents are pathways (views) through the network which determine the subset of parameters that are used and updated by the forwards and backwards passes of the backpropogation algorithm. During learning, a tournament selection genetic algorithm is used to select pathways through the neural network for replication and mutation. Pathway fitness is the performance of that pathway measured according to a cost function. We demonstrate successful transfer learning; fixing the parameters along a path learned on task A and re-evolving a new population of paths for task B, allows task B to be learned faster than it could be learned from scratch or after fine-tuning. Paths evolved on task B re-use parts of the optimal path evolved on task A. Positive transfer was demonstrated for binary MNIST, CIFAR, and SVHN supervised learning classification tasks, and a set of Atari and Labyrinth reinforcement learning tasks, suggesting PathNets have general applicability for neural network training. Finally, PathNet also significantly improves the robustness to hyperparameter choices of a parallel asynchronous reinforcement learning algorithm (A3C). \end{abstract}

%
%

%
%
\printccsdesc



\section{Introduction}
A plausible requirement for artificial general intelligence is that many users will be required to train the same giant neural network on a multitude of tasks. This is the most efficient way for the network to gain experience, because such a network can reuse existing knowledge instead of learning from scratch for each task. To achieve this, we propose that each user of the giant net be given a population of agents whose job it is to learn the user-defined task as efficiently as possible. Agents will learn how best to re-use existing parameters in the environment of the neural network by executing actions within the neural network. They must work in parallel with other agents who are learning other tasks for other users, sharing parameters if transfer is possible, learning to update disjoint parameters if interference is significant. Each agent may itself be controlled by an arbitrarily complex reinforcement learning algorithm, but here we chose the very simplest possible `agent', a unit of evolution \cite{fernando2011evolvable}.\\

The framework for AGI described above includes aspects of transfer learning \cite{taylor2011introduction}, continual learning \cite{ring1994continual} and multitask learning \cite{caruana1998multitask}. Our work shares a motivation with a recent paper ``Outrageously large neural networks" in which the authors write that ``the capacity of a neural network to absorb information is limited by its number of parameters" \cite{outrage}. If a standard neural network is naively trained training cost scales quadratically with model width, whereas PathNet theoretically has constant computation speed with respect to the network width because only a fixed-size subset of the larger network is used for the forwards and backwards pass at any time (although there is no guarantee that more training may not be required in some cases). Our work is related also to ``Convolutional Neural Fabrics" in which connection strengths between modules in the fabric are learned, but where (unlike PathNet) the whole fabric is used all the time \cite{saxena2016convolutional}. \\
 
\begin{figure*}[h]
\centering
\includegraphics[width=170mm]{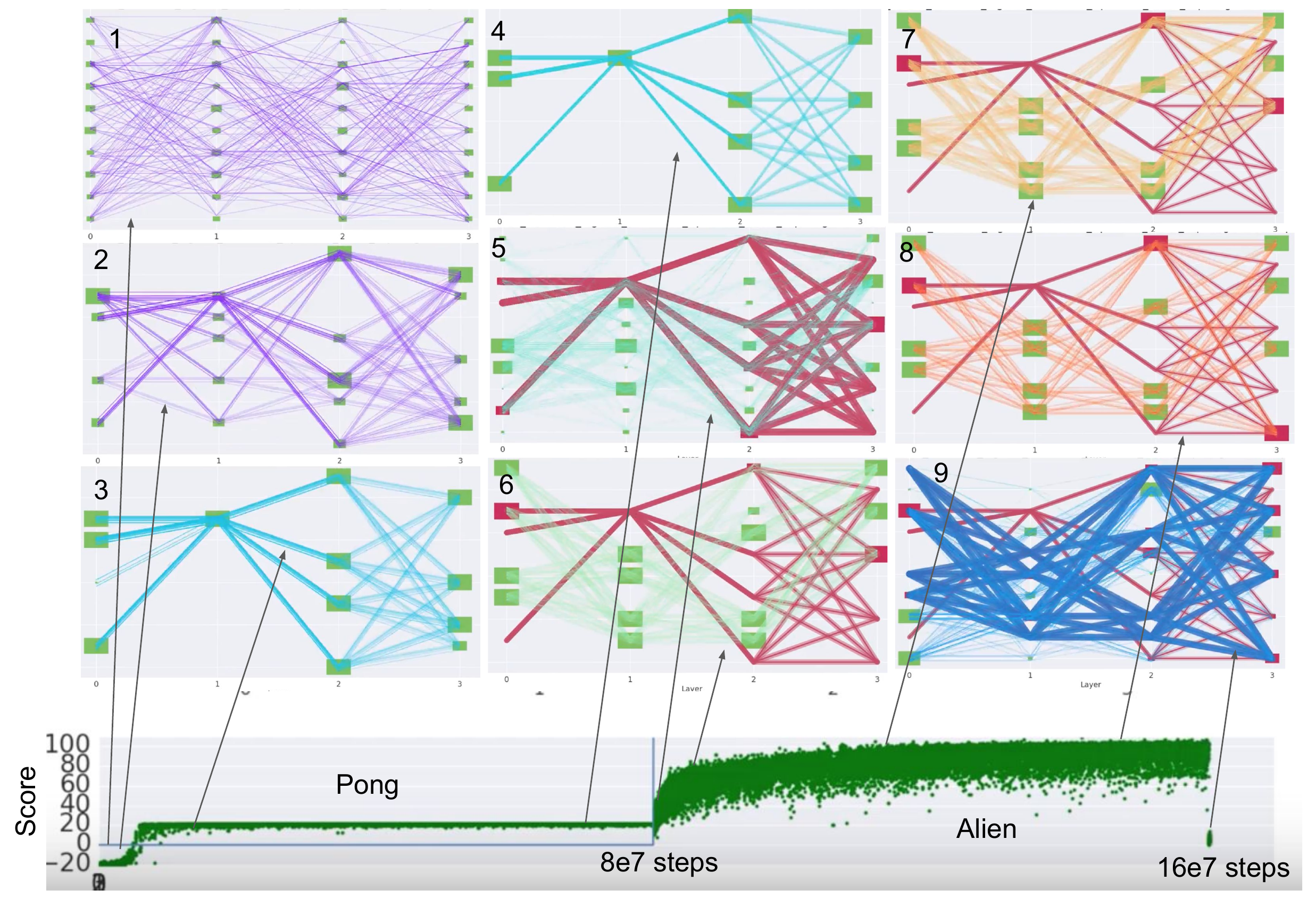}
\caption{A population of randomly initialized pathways (\textcolor{purple}{purple lines} in Box 1) are evolved whilst learning task A, Pong. At the end of training, the best pathway is fixed (\textcolor{darkred}{dark red lines} in Box 5) and a new population of paths are generated (\textcolor{lightblue}{light blue lines} in Box 5) for task B. This population is then trained on Alien and the optimal pathway that is evolved on Alien is subsequently fixed at the end of training, shown as \textcolor{darkblue}{dark blue lines} in Box 9.}
\label{fig:7}
\end{figure*}

This paper introduces PathNet, a novel learning algorithm with explicit support for transfer, continual, and multitask learning. Figure~\ref{fig:7} illustrates the algorithm in action. The first task A is Pong and the second task B is Alien; both are trained consecutively for 80M timesteps each. The \textcolor{purple}{purple lines} in Box 1 of the figure shows all 64 randomly initialized paths through the neural network model at the beginning of Pong training. A tournament selection genetic algorithm is then used to evolve paths, where during a fitness evaluation the path is trained for a few game episodes by gradient descent using a reinforcement learning algorithm. Thus, evolution and learning are taking place simultaneously, with evolution only guiding where gradient descent should be applied to change the weight and bias parameters. Box 2 shows the population converging (with many paths overlapping with each other) as performance improves. As perfect performance is achieved the population converges to a single path as shown in Box 3. Box 4 shows that the converged single pathway persists until the end of the training session. At this point the task switches to task B (Alien) and the optimal path for Pong gets 'fixed', i.e. the modules on that path have their weights and biases frozen. Box 5 shows the fixed path as heavy \textcolor{darkred}{dark red} lines alongside a new randomly initialized population of paths in \textcolor{lightblue}{light blue}. The new population of paths evolve and converge on Alien by Box 8. After 160M steps, the optimal path for Alien was fixed, shown as \textcolor{darkblue}{dark blue lines} in Box 9. \\

PathNets evolve a population of pathways through a neural network that scaffolds and channels any desired gradient-descent-based learning algorithm towards a limited subset of the neural network's parameters and then fixes these parameters after learning so that functionality can never be lost; it resembles progressive neural networks, in that catastrophic forgetting is prevented by design \cite{rusu2016progressive}. In progressive neural networks the topology determining transfer is hard-wired rather than learned, whereby the first neural network is trained on the source task and then a second neural network is trained on the target task which receives inputs from the first neural network, which has its weights fixed. PathNets allow the relationships between the original `columns' and later `columns' to be evolved, where a column is one deep neural network.\\

Two examples of PathNets were investigated, a serial implementation on two supervised learning tasks where the PathNet is trained by stochastic gradient descent, and a parallel implementation on reinforcement learning tasks (Atari and Labyrinth games) where the PathNet is trained by the Async Advantage Actor-Critic (A3C) . A3C is an efficient distributed reinforcement learning algorithm which runs on multiple CPUs with e.g. 64 asynchronously updated workers that simultaneously share and update the parameters of a single network\cite{mnih2016asynchronous}. Positive transfer from a source task to a target task is demonstrated in all four domains, compared to single fixed path controls trained from scratch and after fine-tuning on the first task.\\ 

The concept of the PathNet was first conceived of within the framework of Darwinian Neurodynamics as an attempt to envisage how evolutionary algorithms could be implemented in the brain \cite{fernando2012selectionist}. However, in that original work both the topology and the weights of the path were evolved and there was no gradient descent learning \cite{fernando2011evolvable}. Performance was comparable with a standard genetic algorithm on combinatorial optimization problems. Here we show performance superior to A3C and stochastic gradient descent for transfer learning. \\

\begin{figure*}[h]
\centering
\includegraphics[width=170mm]{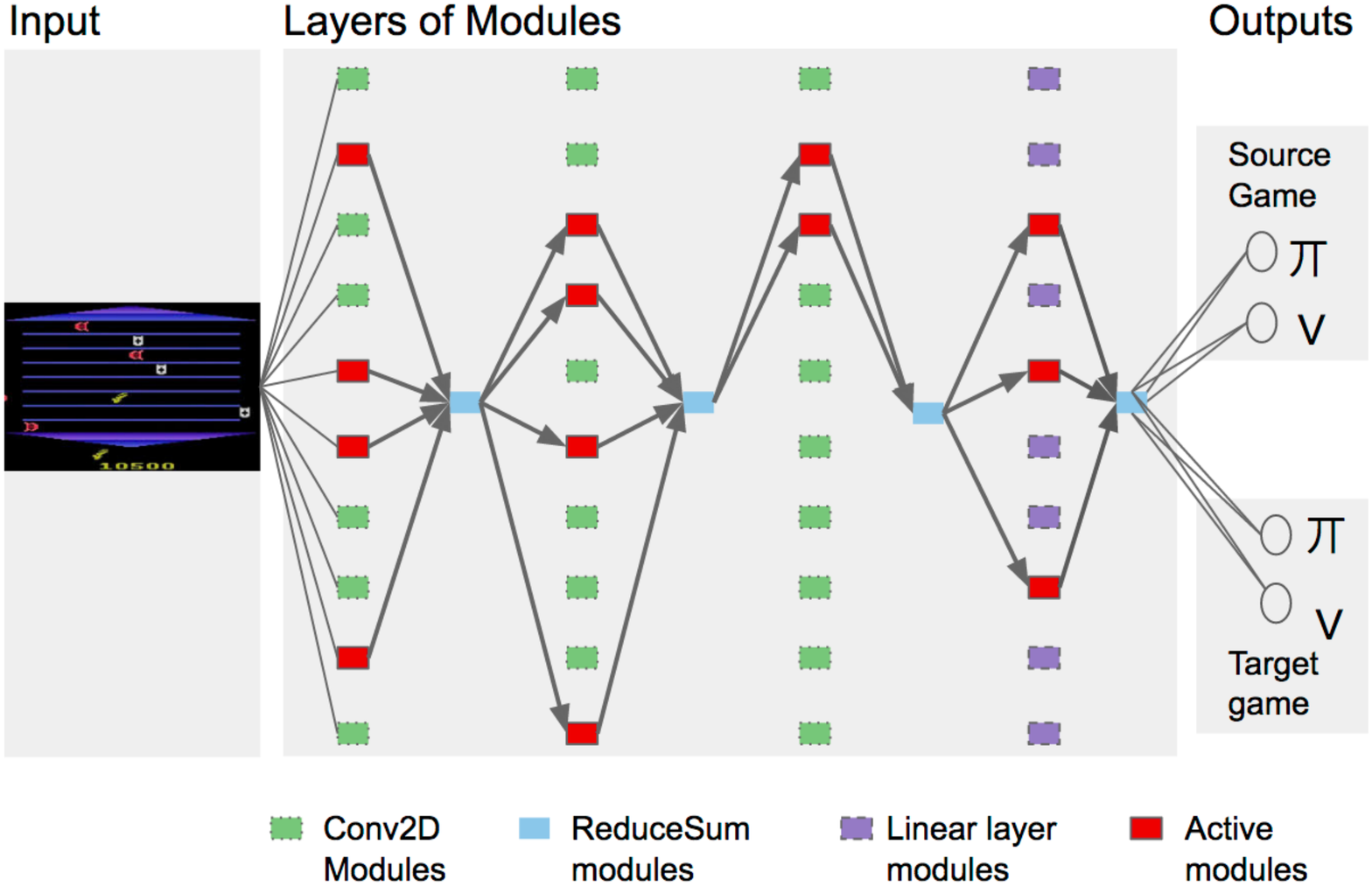}
\caption{The PathNet used for learning Atari and Labyrinth games consists of a 4 layer network with 10 (or sometimes 15) modules in each layer. The first three layers' modules are convolutional 2D kernels with 8 kernels per module (\textcolor{fig2_green}{green boxes} in figure), kernel sizes (8, 4, 3), and strides (4, 2, 1) from 1st to 3rd layers respectively, and with the final layer consisting of fully connected linear layers of 50 neurons each (\textcolor{fig2_purple}{purple boxes}). After each module there is a rectified linear unit. Between layers the feature maps are summed before being passed to the modules of the subsequent layer (\textcolor{fig2_blue}{blue boxes}). Typically a maximum of 4 modules per layer are permitted to be included in a pathway (shown as \textcolor{fig2_red}{red boxes}), otherwise evolution would simply grow the pathway to include the whole network as this would improve fitness by growing the number of parameters that can be tuned by learning.}
\label{fig:1}
\end{figure*}

\begin{figure*}[h!]
\centering
\includegraphics[width=80mm]{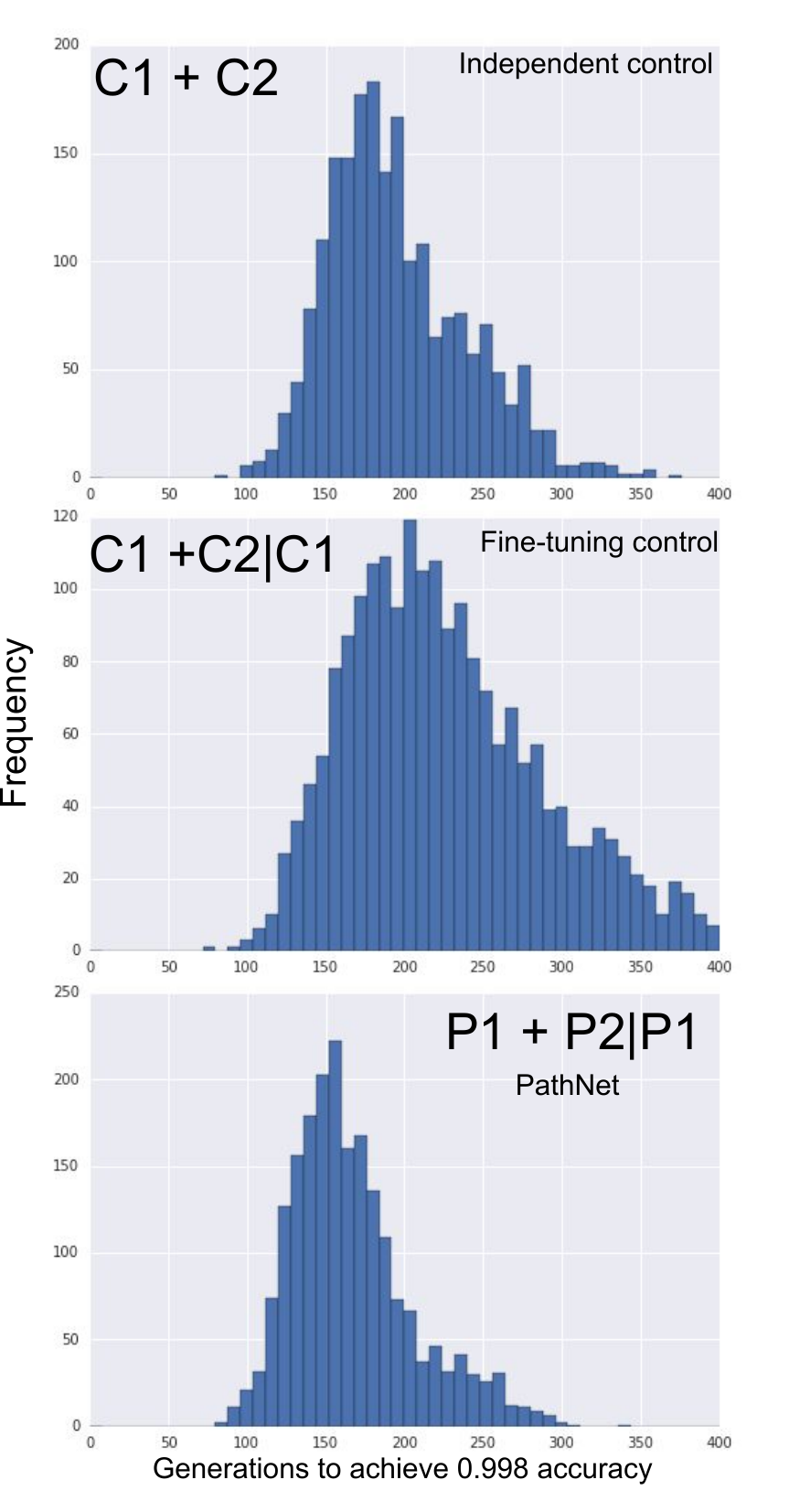}
\caption{PathNet is capable of supporting transfer learning in binary MNIST classification. The top figure shows C1 + C2 which is the sum of learning times over the source and target tasks required to learn both tasks to 0.998 accuracy, with a separate maximum sized fixed path for each task, and thus constitutes the independent learning control; mean = 195 generations. The middle figure shows C1 + C2|C1, which is the time required to learn both tasks where the second task is not learned from scratch, but by fine tuning the same maximum sized fixed path as was used to learn the first task; mean = 229 generations. The bottom graph shows the sum of times required to learn the source and target tasks using PathNet; mean = 167 generations.}
\label{fig:2}
\end{figure*}

\begin{figure*}[h!]
\centering
\includegraphics[width=80mm]{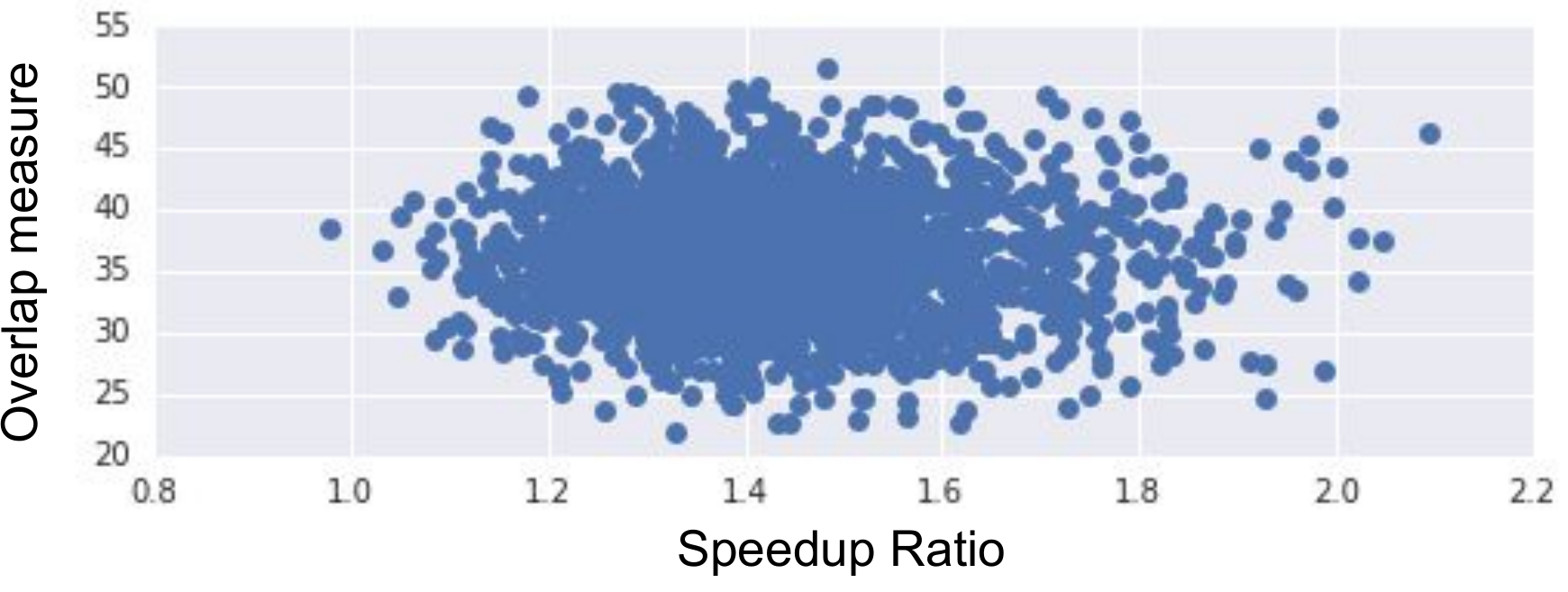}
\caption{Overlap (y-axis) is not correlated with speedup ratio (C1 + C2)/(P1 + P2|P1) suggesting that PathNet discovers how much overlap to use based on whether the tasks could benefit from overlap or not.}
\label{fig:3}
\end{figure*}

\begin{figure*}[h!]
\centering
\includegraphics[width=80mm]{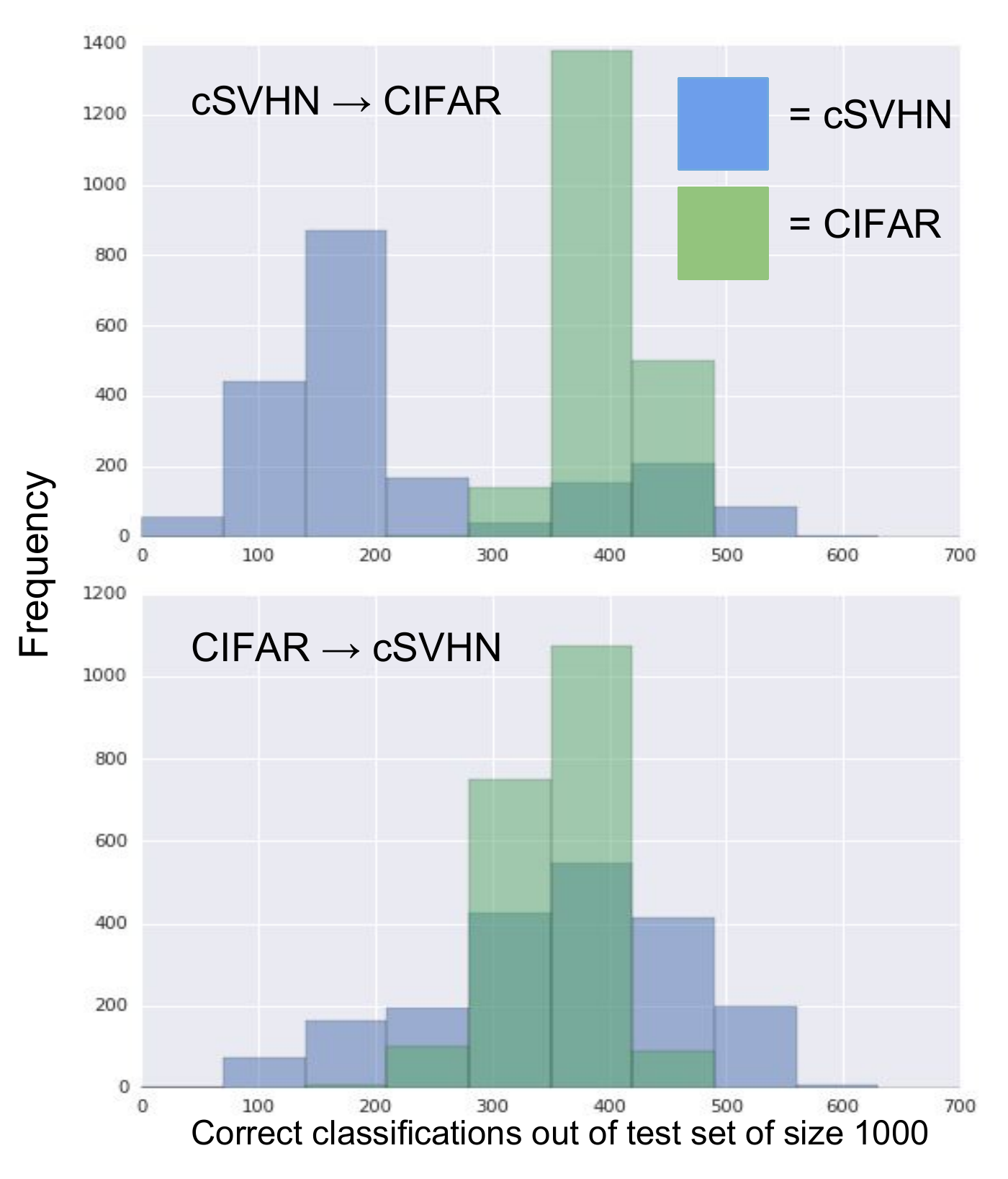}
\caption{Top graph shows a histogram of accuracies when learning cSVHN first and CIFAR second. The bottom graph shows a histogram of accuracies when learning CIFAR first and cSVHN second. Learning either task second with PathNet results in improved performance, suggesting that PathNet is able to use weights in the optimal path found in task A, when learning task B.}
\label{fig:4}
\end{figure*}

\begin{figure*}[h!]
\centering
\includegraphics[width=162mm]{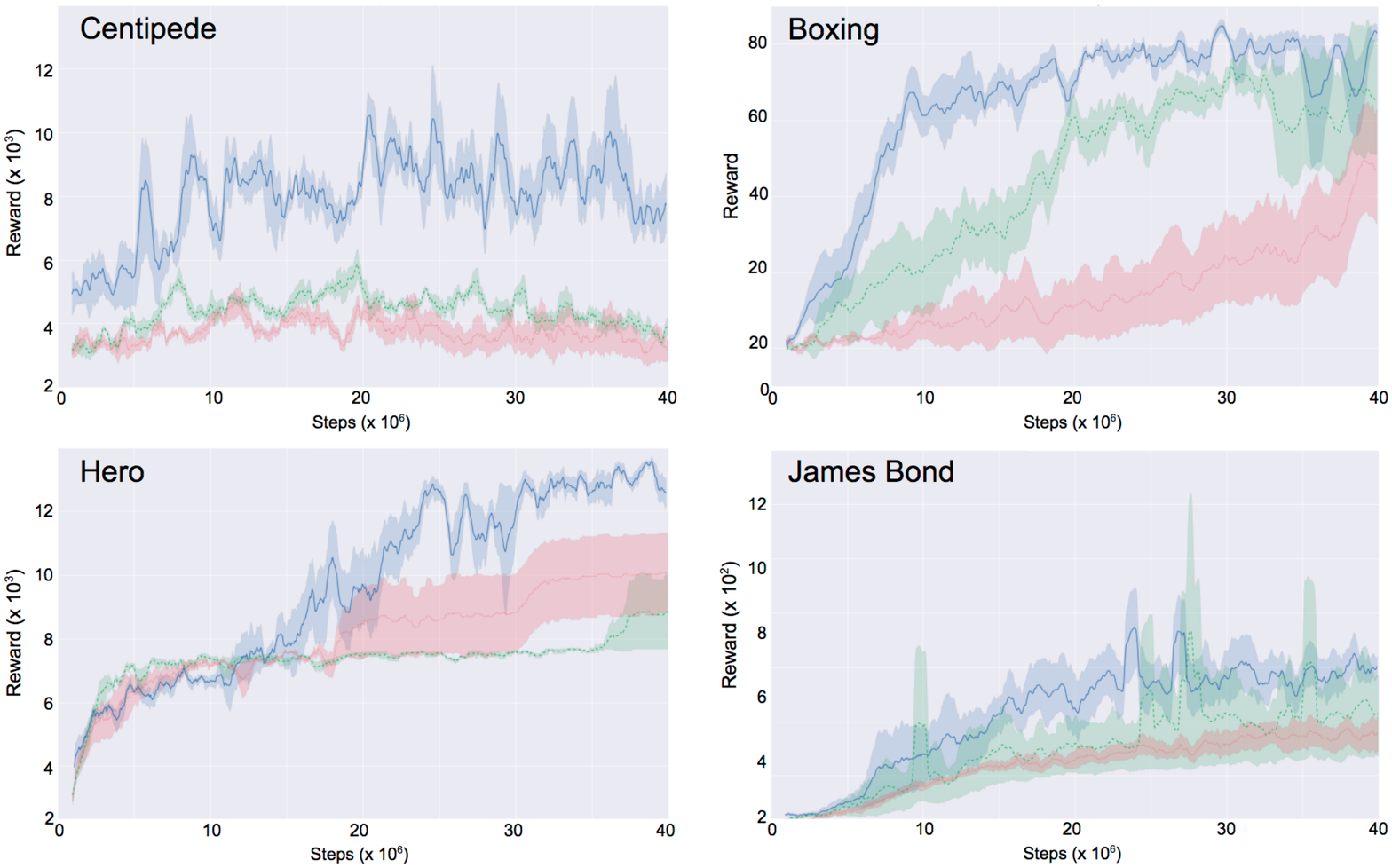}
\caption{PathNet shows positive transfer from River Raid to Centipede, Boxing, Hero and James Bond. The graphs show the reward over the first 40 million steps of training. Blue shows the results from the best five hyperparameter settings of PathNet out of 243. Compare these to the best five hyperparameter setting runs for independent learning (red) and fine-tuning (green) controls out of 45.}
\label{fig:5}
\end{figure*}

\begin{figure*}[h!]
\centering
\includegraphics[width=160mm]{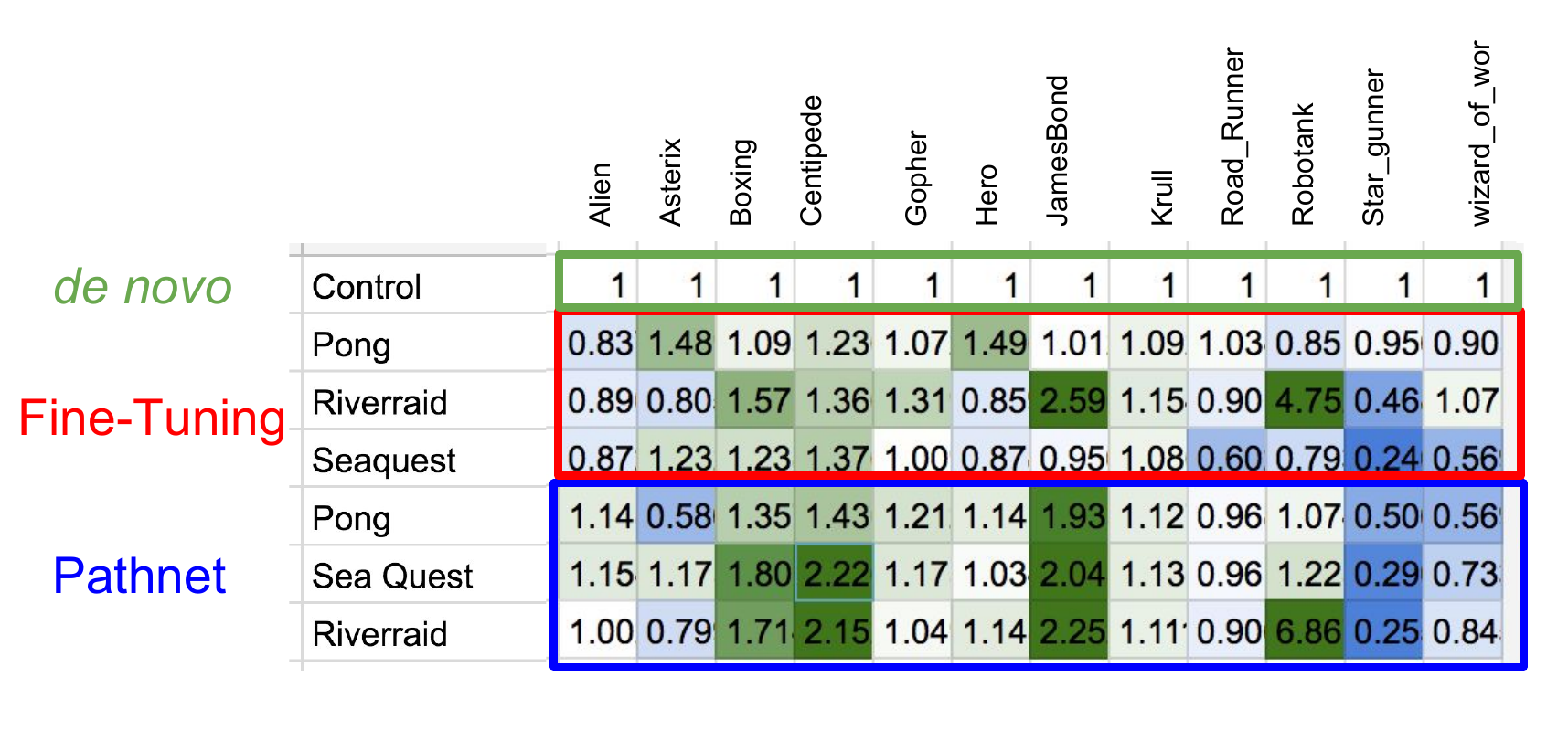}
\caption{We report results by showing the best run from a hyperparameter search. Performance is evaluated by measuring the area under the learning curve (average score per episode during training), rather than final score. The transfer score is then defined as the relative performance of an architecture compared with a independent baseline (control) with a fixed maximum size path, trained only on the target task (top row). This ratio is >1 if there is speedup, and <1 if there is slowdown. We present transfer score curves for selected source-target games, and summarize all such pairs in this transfer matrix. The next 3 rows show fine-tuning controls, and the final three rows show PathNet. Green means positive transfer, and blue means negative transfer.}
\label{fig:mainResult}
\end{figure*}

\begin{figure*}[h!]
\centering
\includegraphics[width=180mm]{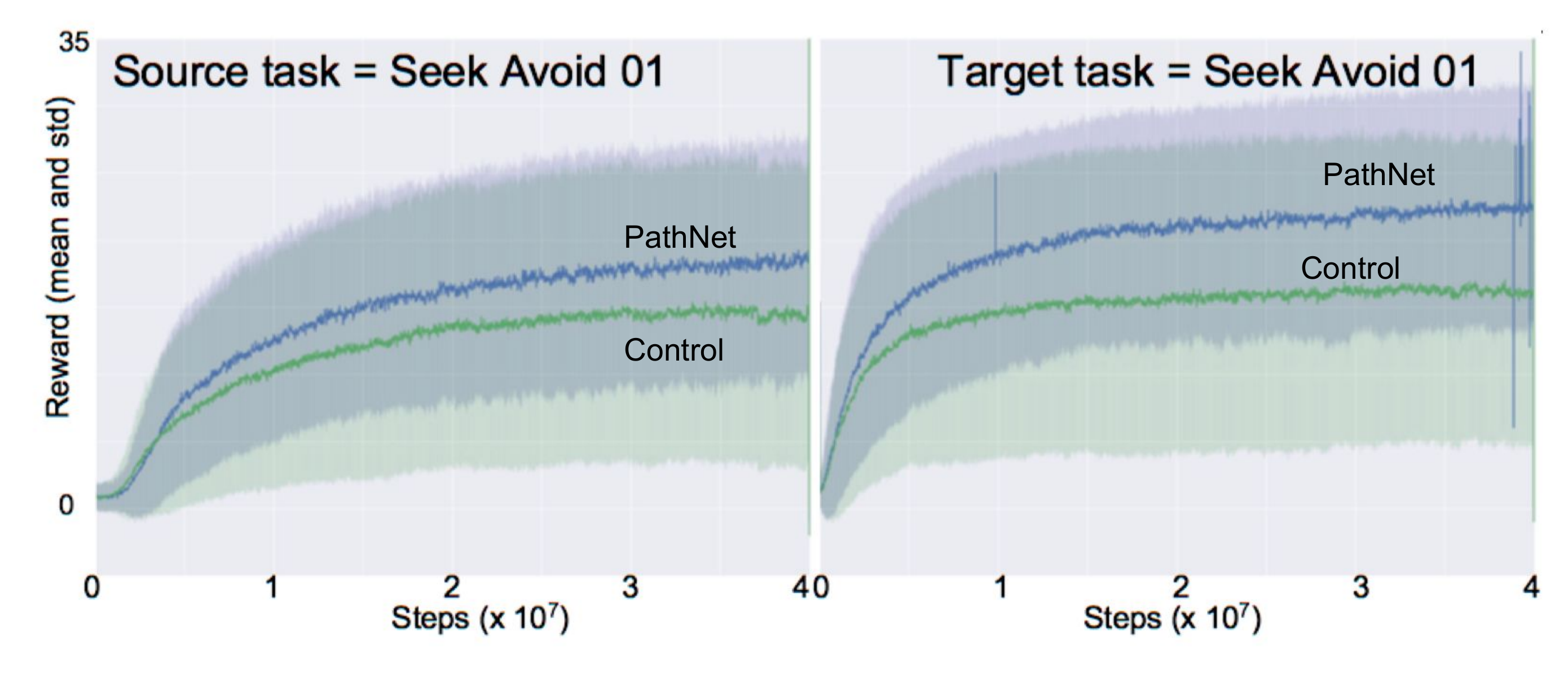}
\caption{PathNet outperforms the de novo control and the fine-tuning control on learning seekavoid\_arena for the first time, on average over all hyperparameters explored, out of 243 experiments. It also outperforms fine tuning when seekavoid\_arena \cite{jaderberg2016reinforcement} is learned a second time, with a new value and policy readout, with the same set of 243 hyperparameter settings.}
\label{fig:robustness}
\end{figure*}

\begin{figure*}[h!]
\centering
\includegraphics[width=180mm]{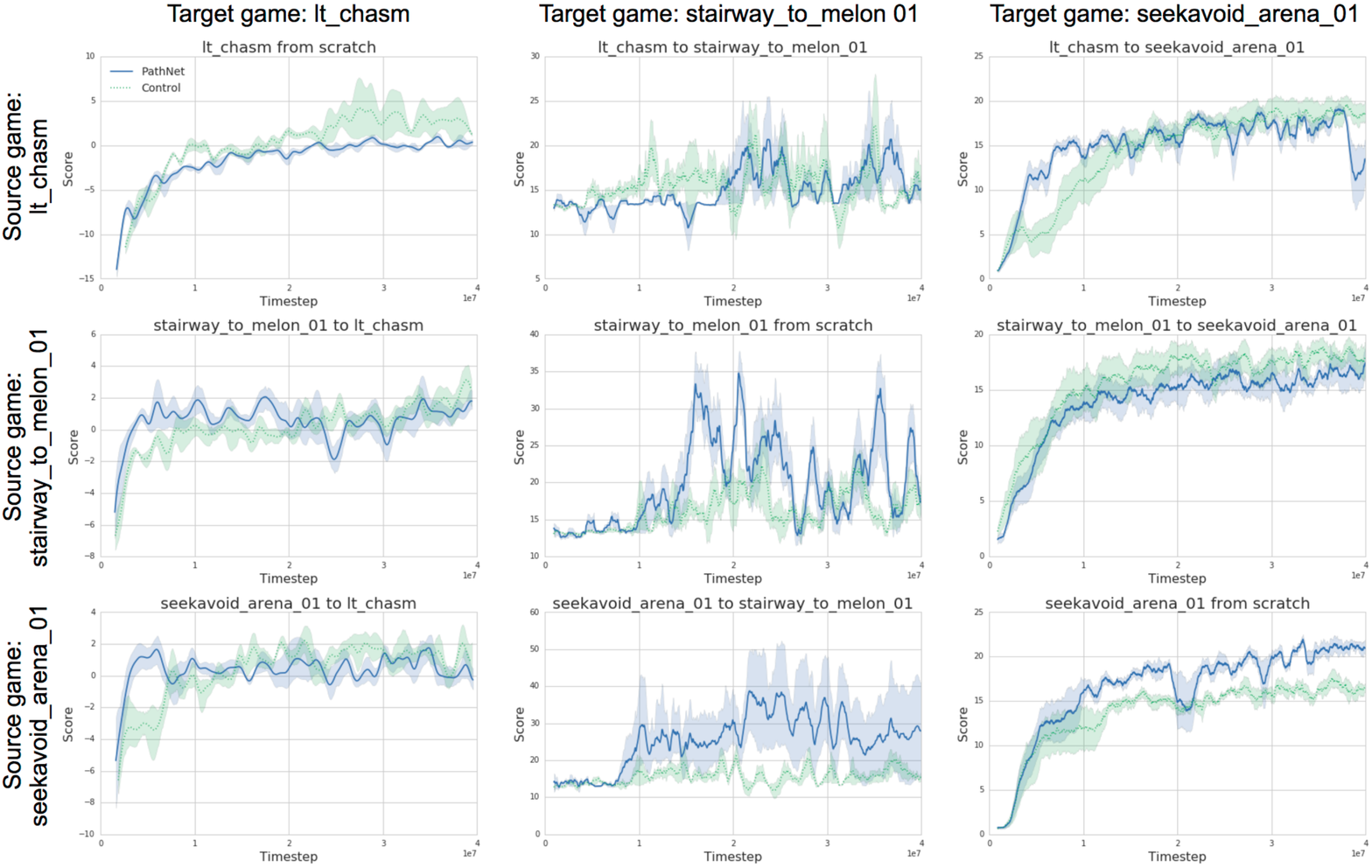}
\caption{Means of the best 5 (out of a size 24 parameter sweep) training runs for PathNet compared with the fine-tuning (off diagonal) and independent path (diagonal) controls for the three Labyrinth levels considered. The diagonal results show the performance of PathNet when learning from scratch compared to the independent fixed path control. Off-diagonal results show the performance of pathNet during transfer. }
\label{fig:transfer}
\end{figure*}

\begin{figure*}[h!]
\centering
\includegraphics[width=160mm]{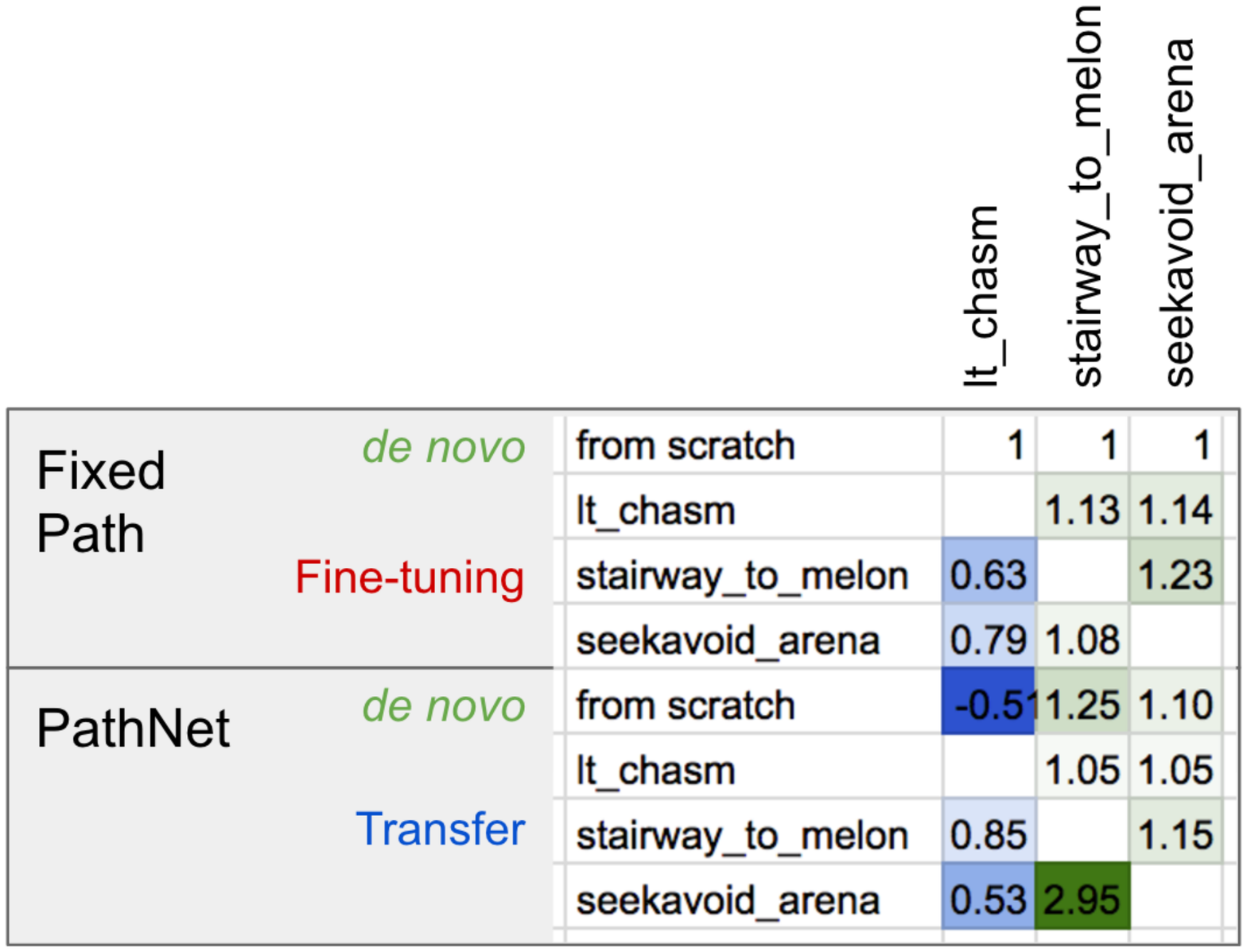}
\caption{Results of the best runs from a hyperparameter search. The numbers in the table show the relative performance of an architecture learning a target task compared with an independent baseline with a fixed maximum size path trained from scratch only on the target task. A number >1 (green) represents improved performance and <1 (blue) is worse performance. The top row ('Fixed path de novo from scratch') are the controls, i.e. the performance of the fixed-path network learning the target task (each column) from scratch. The rows below show fixed-path performance on fine-tuning and PathNet, learning de novo from scratch, and PathNet performance transferring from task A to task B.}
\label{fig:transferTable}
\end{figure*}

\begin{figure*}[h!]
\centering
\includegraphics[width=180mm]{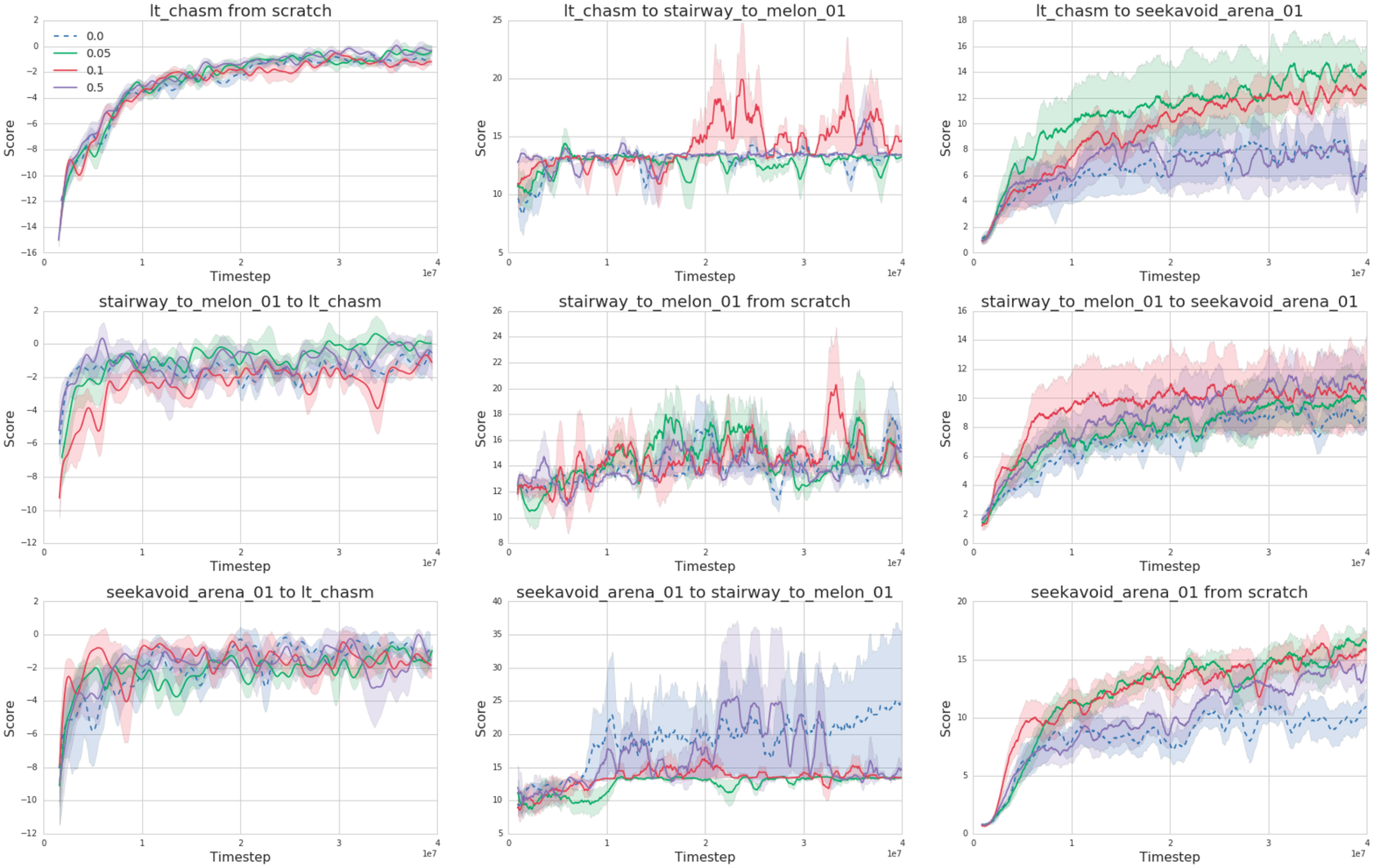}
\caption{The contribution of Module Duplication on performance. Graphs show the mean performance across all hyperparameters for PathNet with different module duplication rates [0, 0.05. 0.1, 0.5] per episode completed by worker 0. For some tasks (e.g. learning seekavoid\_arena module duplication has been beneficial.}
\label{fig:transferNet2Net}
\end{figure*}

\begin{figure*}[h!]
\centering
\includegraphics[width=162mm]{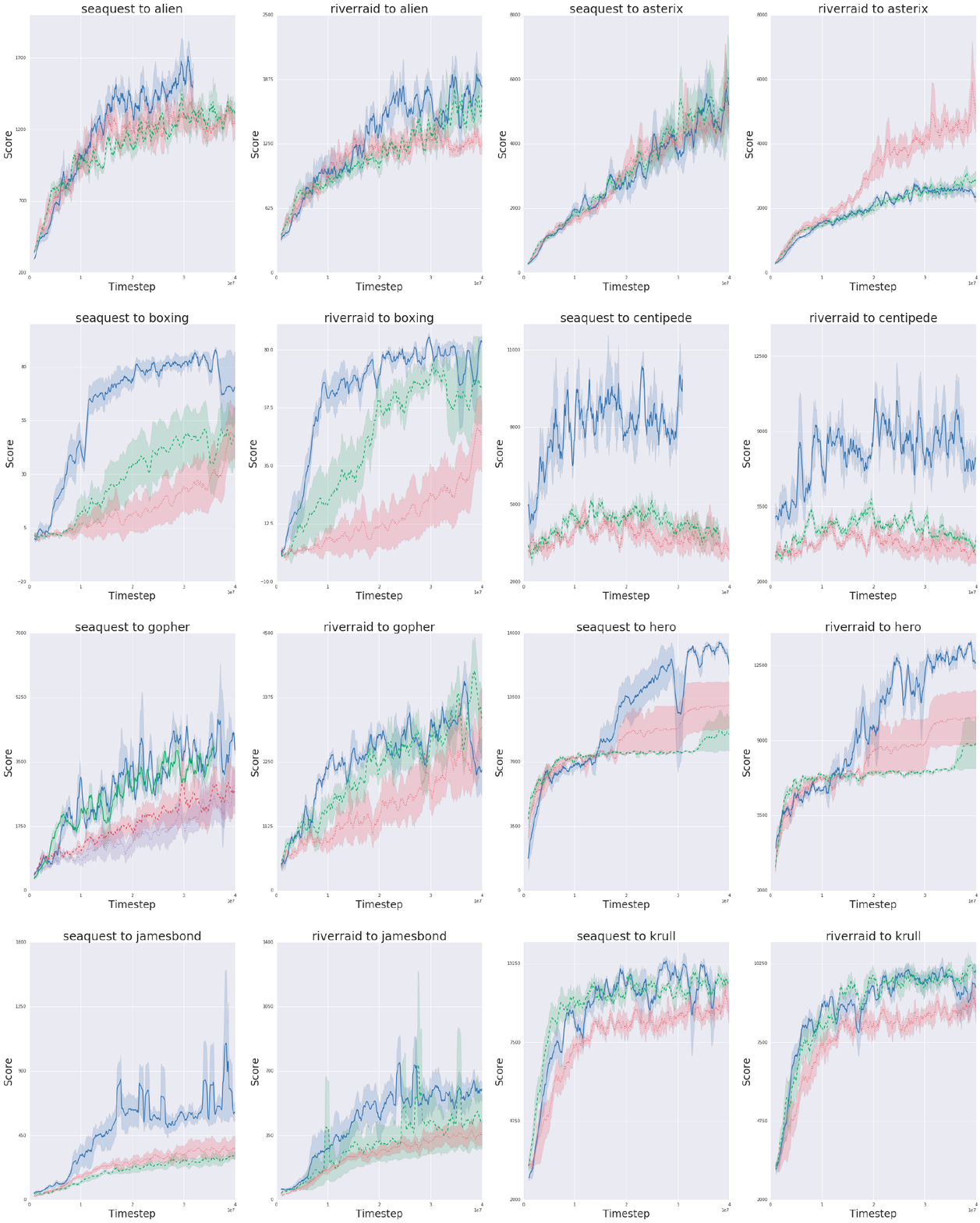}
\caption{Transfer results on various Atari games. The graphs show the reward over the first 40 million steps of training. Blue shows the results from the best five hyperparameter settings of PathNet out of 243. Compare these to the best five hyperparameter setting runs for independent learning (red) and fine-tuning (green) controls out of 45.}
\label{fig:atariGrid}
\end{figure*}

\section{Methods} 

\subsection{PathNet Architecture} 
A PathNet is a modular deep neural network having \(L\) layers with each layer consisting of \(M\) modules. Each module is itself a neural network, here either convolutional or linear, followed by a transfer function; rectified linear units are used here. For each layer the outputs of the modules in that layer are summed before being passed into the active modules of the next layer. A module is \(active\) if it is present in the path genotype currently being evaluated (see below). A maximum of \(N\) distinct modules per layer are permitted in a pathway (typically \(N = 3\) or \(4\)). The final layer is unique and unshared for each task being learned. In the supervised case this is a single linear layer for each task, and in the A3C case each task (e.g. Atari game) has a value function readout and a  policy readout (see \cite{mnih2016asynchronous} for a complete description of the A3C algorithm used). \\

\subsection{Pathway Evolution: Serial and Parallel} 
\(P\) genotypes (pathways) are initialized randomly, each genotype is at most a \(N\) by \(L\) matrix of integers, which describe the active modules in each layer in that pathway. In the serial supervised implementation, a binary tournament selection algorithm is implemented in series as follows. A random genotype is chosen, and its pathway is trained for \(T\) epochs, its fitness being the negative classification error during that period of training. Then another random genotype is chosen and its pathway trained for T epochs. A copy of the winning pathway's genotype overwrites the losing pathways genotype. The copy of the winning pathway's genotype is then mutated by choosing independently each element with a probability of \(1/[N\times L]\) and adding an integer in the range \([-2, 2]\) to it.  A local neighbourhood was used to promote spatial localization of network functionality. \\

In the A3C (reinforcement learning) case, all 64 genotypes are evaluated in parallel, one by each of the 64 workers. Therefore pathways restrict the simultaneous updates of parameters by the workers to only their specific subsets, as opposed to the standard A3C algorithm in which all workers update all parameters. The fitness of a genotype is the return accumulated over the \(T\) episodes that a worker played using that genotypes pathway. While a worker is evaluating, it writes a large negative fitness to the shared fitness array, so that no genotypes wins a tournament until it has been evaluated. Once the worker has finished \(T\) episodes, it chooses \(B\) other random genotypes and checks if any of those genotypes have returned a fitness of at least its own fitness. If at least one has, then the highest fit genotype overwrites the current worker's genotype, and is mutated as above.  If no other worker had a genotype with fitness greater than this workers own genotype, then the worker reevaluates its own genotype. 

\subsection{Transfer Learning Paradigm} 
Once task A has been trained for a fixed period of time or until some error threshold has been reached, the best fit pathway is fixed, which means its parameters are no longer allowed to change. All other parameters not in an optimal path are reinitialized. We found that without re-initialization transfer performance did not exceed that of fine-tuning. In the A3C case (but not the supervised learning cases) the original best fit pathway is always \(active\) during the forwards pass of the network, in addition to the newly evolving pathway, but its parameters are not modified by the backwards pass. A new set of random pathways is then initialized and evolved/trained on task B. In both the supervised and reinforcement settings, pathNet is compared with two alternative setups: an independent learning control where the target task is learned de novo, and a fine-tuning control where the second task is learned with the same path that learned the first task (but with a new value function and policy readout).  \\

\subsection{Binary MNIST classification tasks} 
A binary MNIST classification involves distinguishing two classes of MNIST digits from one another, for example 5 verses 6 \cite{lecun1998gradient}. To make the task more difficult, salt and pepper noise of 50\% is added to the MNIST digits. A transfer experiment involves training and evolving paths on the first task until perfect classification on the training set is achieved.\\

At this point, a new population of path genotypes is initialized and evolved on the second task until perfect performance on the training set is achieved. Between tasks the following modifications are made to the underlying network. The parameters contained in the optimal path evolved on the first task are fixed, and all other parameters are reset to their random initial values. The reported outcome measures are the training times required to reach this classification accuracy. The overall PathNet consists of \(L = 3\) layers. Each layer contains \(M = 10\) linear units with 20 neurons each followed by rectified linear units. Between layers the activations of the units are summed. Reading out from the final layer is a separate linear layer for each task.\\

A maximum of 3 of these units per layer can be included in a single pathway. The genotype describing this pathway is a \(3 \times 3\) matrix of integers in the range \([1,10]\). A population of 64 of these genotypes was generated uniformly randomly at the start of each task. The overall algorithm is as follows. Two paths are chosen randomly for evaluation. The evaluation of one path involves training with stochastic gradient descent with learning rate 0.0001 on 50 mini-batches of size 16. The fitness of that pathway is the proportion of correct examples classified on the training set during this training period. Once the fitness of both pathways has been calculated, the pathway with the lower fitness is overwritten by a copy of the pathway with the higher fitness and mutated. Mutation takes place with equal probability \(1/(3 \times 3) \) per element of the genotype, adding a new random integer from range\([-2, 2]\) to the current value of that element. This is a binary tournament selection algorithm \((B = 2)\) \cite{harvey2011microbial}.\\ 

\subsection{CIFAR and SVHN classification tasks} 
The larger version of the above network is used to train on CIFAR and cropped SVHN \cite{netzer2011reading} of standard size \(28 \times 28\) with\( L = 3\) and \(M = 20\) modules per layer of 20 neurons each, and with pathways that may contain up to 5 modules per layer. In this case the networks were not trained to some fixed accuracy, but for a fixed period of 500 generations. Each generation consisted of the evaluation of two pathways, each for 50 mini-batches of size 16 images. The performance measure here was the accuracy achieved after this fixed training time. Evidence for positive transfer in this case is given by the second task showing a higher final accuracy compared to when it was learned from scratch.\\

\subsection{Atari games} 
 We tested whether there was a speedup in learning a second (target) game after having played either Pong, RiverRaid or Seaquest as a source game. The target games were as follows: Alien, Asterix, Boxing, Centipede, Gopher, Hero, JamesBond, Krull, RoadRunner, StarGunner, WizardofWor. These are the same games presented by the authors of Progressive Neural Networks. In this case the A3C algorithm was used with 64 workers running in parallel. The first game is learned for 80M (or 40M) timesteps of training in total across all workers. During this time, workers are evolving pathways through the network. After the first game is learned the population of paths is reinitialized and the workers play the second game for 40M timesteps in total. Between tasks the following modifications are made to the underlying network. The parameters contained in the optimal path evolved on the first task are fixed, and all other parameters are reset to their random initial values. Evolution takes place according to a distributed asynchronous tournament selection algorithm which works as follows. The population of pathways and their fitnesses is stored in a central parameter server. Each worker accesses its own particular index into this population of genotypes corresponding to that worker's id. While it is first evaluating that pathway it writes -1000 to the fitness array at a position corresponding to it's id. After \(T = 10\) episodes that worker writes the total sum of rewards obtained into the fitness array. It then observes the fitnesses of \(B = 20\) other random workers and if any worker has a higher fitness than itself, it chooses the highest fitness worker in that set, and copies the pathway genotype to its own location with mutation, and resets its own fitness to -1000. However, if no other worker of the \(B\) chosen has a path better than its own, then its own pathway is re-evaluated without resetting the fitness obtained from the previous evaluation. Mutation is as before, with the minor modification that if a location is chosen for mutation, then there is a 20\% chance (during the second task) that a module from the optimal pathway evolved in the first task will be incorporated into the genotype. In the variant of PathNet used for this task, the optimal path evolved in the first task is always active during the forwards pass, in this sense, this version resembles progressive nets more closely than the supervised versions presented above. The architecture of the network used consists of a core of \(L = 4\) layers, each with a width of \(M = 10\) modules: namely, 3 convolutional layers where each module consists of 8 kernels, and a final layer of fully connected modules of 50 hidden nodes each. Between layers the feature maps and activations are summed. Reading out from the final layer are two linear layers for each game, one encoding a value function and the other encoding an action policy. Figure~\ref{fig:1} shows the network architecture: The first three layers' modules are shown as \textcolor{fig2_green}{green boxes}; final layer of modules shown as \textcolor{fig2_purple}{purple boxes}; the between-layer summing modules are shown as \textcolor{fig2_blue}{blue boxes}; and active modules specified by the pathway as \textcolor{fig2_red}{red boxes}. Readout units not involved in the PathNet itself are shown as circles on the right. \\

\subsection{Labyrinth Games}
We investigated the performance of a very similar PathNet architecture on 3 Labyrinth games. Labyrinth is a 3D first person game environment \cite{jaderberg2016reinforcement}. We used the same settings used for Atari games. We chose three games: 'Laser Tag Chasm' (lt\_chasm), 'Seek Avoid Arena 01' (seekavoid\_arena) and 'Stairway to Melon 01' (stairway\_to\_melon). lt\_chasm takes place in a square room with four opponents that must be tagged for points. A chasm in the center of the room must be avoided and shield pick-ups exist that the player must jump to reach. seekavoid\_arena is a 3D room containing apples and lemons. The player must pick up the apples whilst avoiding the lemons. stairway\_to\_melon offers the player two options; the player can either collect a set of small rewards (apples) or they can opt to take a punishment (lemon) in order to reach a melon which results in a higher net reward. 

The PathNet architecture used for Labyrinth was identical to the one used for Atari with the exception of a module-duplication mechanism. In the previous models there has not been any actual copying of parameters, only copying of views over a fixed set of parameters. In the Labyrinth model we enabled PathNet to copy the weights of modules to other modules within the same layer, emulating Net2Net \cite{chen2015net2net} or a distillation \cite{hinton2015distilling} like operation. To aid this, we measure the extent to which a module contributes to the fitness of all the paths it is in by taking a sliding mean over the fitness of any path which contains that module. Using this measure it is possible to bias a weight-copying operator such that currently more globally useful modules are more likely to be copied across into other modules and hence into other paths. The hyperparameter \textit{module duplication rate} determines the rate at which modules would be allowed to duplicate themselves. For the Labyrinth games both the first and second tasks were trained for 40M timesteps. \\ 

\section{Results} 

\subsection{Binary MNIST Classification} 

Figure~\ref{fig:2} shows that with PathNet, learning a source MNIST classification task helps speed up learning in the target MNIST classification task (mean time to solution = 167 generations); greater than the speedup achievable by fine-tuning a fixed path that had initially been trained on the source task (mean time to solution = 229), and also compared to de novo learning of the target task from scratch (mean time to solution = 195). PathNet (bottom graph) learns in fewer generations with less data than both the fine-tuning (middle) and independent learning controls (top). The control networks consisted of exactly the same learning algorithm, but with no evolution of paths, and only one fixed maximum size pathway through the network. The total speedup ratio compared to the independent controls was 1.18. \\

Videos showing performance of PathNet can be obtained online at https://goo.gl/oVHMJo. They reveal that the modules in early layers of the big network converge in the population of pathways the quickest, followed by later layers, there being much more exploration and training of all the modules in the final layer. Many modules in the final layer contribute to high fitness, whereas only a few modules in the first layer do so. Thus, a population provides an elegant solution to the exploration /exploitation trade-off in a layer specific manner. Analysis did not reveal that the speedup ratio was correlated with path overlap as measured by the number of modules in the original optimal path that were present in the population of paths at the end of the second task. This suggests that speedup can be obtained by PathNet both determining when there should be overlap and when there should not be overlap. The fact that on average fine-tuning is slower than independent task learning in MNIST binary classification tasks implies that generally overlap would not be expected to provide speedup. Thus, PathNet can be seen to have done its job by controlling the amount of overlap properly, see Figure~\ref{fig:3} \\

\subsection{CIFAR and SVHN} 
In this experiment we compare only the accuracy PathNet obtains after a short fixed number of 500 generations. A fully connected network of this size is generally insufficient to perform well on these datasets. After 500 generations, when learning from scratch with PathNet, cSVHN and CIFAR are learned to 25.5\% and 35.3\% accuracy on average, see Figure~\ref{fig:4}. But after the first task has been learned to this accuracy, learning the second task is faster, so when cSVHN and CIFAR are learned as the second task with PathNet, then accuracies of 35.7\% and 39.8\% are achieved respectively. Thus, both cSVHN and CIFAR are learned faster with PathNet when learned second rather than first. 

\subsection{Atari Games} 
Transfer performance on PathNet is compared with fixed maximum-size-path de novo training and fine-tuning controls for which a hyperparameter sweep was conducted involving learning rates [0.001,0.0005,0.0001] and entropy costs [0.01, 0.001, 0.0001]. The following hyperparameters were investigated for PathNet: evaluation time T [1,10,50], mutation rate [0.1, 0.01, 0.001] and tournament size B [2, 10, 20].   

In Figure~\ref{fig:5} the performance of PathNet (blue) is compared with independent and fine-tuning controls (green) on 4 target games, having learned RiverRaid first. In all cases the top 5 runs of a hyperparameter search are shown. We generally found that strong selection with tournament sizes of \(B = 10\), \(T = 10\) game episodes per evaluation, and low mutation rates of 0.01-0.001 were optimal, allowing rapid convergence of paths to a single path, followed by exploration of small variants to a path, thus focusing learning on a few parameters, with occasional exploration of novel bypasses. PathNet is superior to controls in these 4 cases.  \\

Figure~\ref{fig:mainResult} shows that PathNet is on average superior to independent learning and fine-tuning over the Atari games we investigated. Compared to a control of 1.0, fine tuning achieves a 1.16 speedup on average whereas PathNet achieves a 1.33 times speedup. Results for transfer on more Atari games can be seen in Figure~\ref{fig:atariGrid}. \\

\subsection{Labyrinth Games}

PathNet transfer between the three labyrinth games lt\_chasm, seekavoid\_arena and stairway\_to\_melon is compared with fixed maximum-size-path de novo training and fine-tuning controls for which a hyperparameter sweep was conducted involving mutation rates [ 0.1, 0.01, 0.001], module duplication rate [ 0.0, 0.05, 0.1, 0.5] (per episode completed by worker 0) and tournament size B [2, 10]. The learning rate was fixed at 0.001, entropy cost at 0.0001 and evaluation time \(T\) at 13.\\

Figure~\ref{fig:transfer} shows the mean of the training performance for the top 5 experiments for both PathNet and the fixed-path controls. Each row represents a different source game and each column a different target game. Where the source game and target game are the same the graph shows the results of learning the game de novo from scratch. Figure~\ref{fig:transferNet2Net} shows that in some cases the module duplication operation produces improved performance compared to standard PathNet. \\

In several cases (transfer to lt\_chasm and transfer from lt\_chasm to seekavoid\_arena) PathNet learns the second task faster than fine-tuning. Interestingly, PathNet also performs better than fine-tuning when learning stairway\_to\_melon and seekavoid\_arena from scratch.\\

Results of the best runs from a hyperparameter search are summarized in Figure~\ref{fig:transferTable}. Here performance is evaluated by measuring the area under the learning curve (average score per episode during training), rather than final score. The numbers in the table show the relative performance of an architecture learning a target task (each column) compared with an independent baseline with a fixed maximum size path trained from scratch only on the target task. The controls are labelled as 'Fixed path de novo from scratch' (top row) and is 1 for each column (target task). A ratio in a column >1 represents the speedup when learning that column's target task, and <1 is a slowdown. The three rows below the de novo control show the fine-tuning results between the three games. The first row in the PathNet results show performance learning the individual games from scratch and the three rows below that show the PathNet transfer results between the three games.\\

On transferring to lt\_chasm both fine-tuning and PathNet perform worse than the control de novo learning. On the the other two games both exhibit positive transfer performance. The average performance ratio for fine-tuning for transfer across all the game combinations is 1.00 (2 d.p.), essentially no faster than learning from scratch. The average performance ratio for PathNet is 1.26 (2 d.p.); largely due to the good performance transferring from seekavoid\_arena to stairway\_to\_melon.\\

We also compared PathNet to independent and fine-tuning controls over the same sweep of 243 hyperparameters as used in the Atari experiments described above. On the Labyrinth level seekavoid\_arena in which the agent must collect apples but avoid lemons we found that the PathNet had significantly higher mean performance than control runs, both when learning seekavoid\_arena from scratch compared to the de novo controls, and for relearning from the same task, compared to fine-tuning from a network that had previously learned seekavoid\_arena, see Figure~\ref{fig:robustness}. \\

\section{Conclusion} 
PathNet extends our original work on the Path Evolution Algorithm \cite{fernando2011evolvable} to Deep Learning whereby the weights and biases of the network are learned by gradient descent, but evolution determines which subset of parameters is to be trained. We have shown that PathNet is capable of sustaining transfer learning on at least four tasks in both the supervised and reinforcement learning settings.\\

PathNet may be thought of as implementing a form of `evolutionary dropout' in which instead of randomly dropping out units and their connections, dropout samples or `thinned networks' are evolved \cite{srivastava2014dropout}. PathNet has the added advantage that dropout frequency is emergent, because the population converges faster at the early layers of the network than in the later layers. PathNet also resembles `evolutionary swapout' \cite{singh2016swapout}, in fact we have experimented with having standard linear modules, skip modules and residual modules in the same layer and found that path evolution was capable of discovering effective structures within this diverse network. PathNet is related also to recent work on convolutional neural fabrics,  but there the whole network is always used and so the principle cannot scale to giant networks \cite{saxena2016convolutional}. Other approaches to combining evolution and learning have involved parameter copying, whereas there is no such copying in the current implementation of PathNet \cite{auerbach2014online}\cite{chen2015net2net}. \\

Whilst we have only demonstrated PathNet in a fairly small network, the principle can scale to much larger neural networks with more efficient implementations of pathway gating. This will allow extension to multiple tasks. We also wish to try PathNet on other RL tasks which may be more suitable for transfer learning than Atari, for example continuous robotic control problems. Further investigation is required to understand the extent to which PathNet may be superior to using fixed paths. Firstly, a possibility is that mutable paths provide a more useful form of diverse exploration in RL tasks \cite{osband2016deep}. Secondly, it is possible that a larger number of workers can be used in A3C because if each worker can determine which parameters to update, there may be selection for pathways that do not interfere with each other.\\

We are still investigating the potential benefits of module duplication. see Supplementary Video https://goo.gl/oVHMJo. Using this measure it is possible to bias the mutation operator such that currently more globally useful modules are more likely to be slotted into other paths. Further work is also to be carried out in multi-task learning which has not yet been addressed in this paper. \\

Finally, it is always possible and sometimes desirable to replace evolutionary variation operators with variation operators learned by reinforcement learning. A tournament selection algorithm with mutation is only the simplest way to achieve adaptive paths. It is clear that more sophisticated methods such as policy gradient methods may be used to learn the distribution of pathways as a function of the long term returns obtained by paths, and as a function of a task description input. This may be done through a softer form of gating than used by PathNet here. Furthermore, a population (ensemble) of soft gate matrices may be maintained and an RL algorithm may be permitted to 'mutate' these values. \\

The operations of PathNet resemble those of the Basal Ganglia, which we propose determines which subsets of the cortex are to be active and trainable as a function of goal/sub-goal signals from the prefrontal cortex, a hypothesis related to others in the literature \cite{hazy2007towards} \cite{o2006making} \cite{frank2001interactions}. \\

\section{Acknowledgments}
Thanks to Hubert Soyer, Arka Pal, Gabriel Dulac-Arnold, Gabriel Barth-Maron, Matteo Hessel, Alban Rustrani, Stephen Gaffney, Joel Leibo, Eors Szathmary  

%
\bibliographystyle{abbrv}
\bibliography{sigproc}  
%
%

\end{document}